\documentclass{article}
\usepackage{ijcai16}

\usepackage{times}

\usepackage{latexsym}

\usepackage{courier}            
\usepackage[scaled]{helvet} 
\usepackage{url}                  
\usepackage{listings}          
\usepackage{paralist}
\usepackage{enumitem}      

\usepackage[normalem]{ulem}

\usepackage[ruled,vlined,linesnumbered]{algorithm2e}
\SetAlFnt{\footnotesize\sf} 

\usepackage{graphicx}
\usepackage{amsmath,amsfonts,amsthm,amssymb,pgf,tikz}
\usepackage{url}
\usepackage{xspace}
\usepackage{wrapfig}
\usepackage{appendix}
\usepackage{booktabs}
\usepackage{stackengine}

\newcommand{\hide}[1]{}

\newcommand{\q}{\ensuremath{\mathbf{Q}}\xspace}
\newcommand{\evi}{\ensuremath{\mathbf{E}}\xspace}
\newcommand{\m}{\ensuremath{\mathbf{M}}\xspace}

\newcommand{\ul}{\underline{ }}

\newcommand{\bno}{\textbf{-}}
\newcommand{\name}{Swift\xspace}
\newcommand{\optname}{FIDS\xspace}

\makeatletter
\newcounter{transformation}

\makeatother

\newif\ifminted
\mintedtrue 

\title{Swift: Compiled Inference for Probabilistic Programming Languages}
\author{Yi Wu \\
UC Berkeley  \\
jxwuyi@gmail.com
\And
\hspace{-0.5cm}
Lei Li\\
\hspace{-0.5cm}Toutiao.com\\
\hspace{-0.5cm}lileicc@gmail.com
\And
\hspace{-0.5cm}Stuart Russell\\
\hspace{-0.5cm}UC Berkeley\\
\hspace{-0.5cm}russell@cs.berkeley.edu
\And
\hspace{-0.5cm}Rastislav Bodik\\
\hspace{-0.5cm}University of Washington\\
\hspace{-0.5cm}bodik@cs.washington.edu
}

\begin{document}

\maketitle

\begin{abstract}
A probabilistic program defines a probability measure over its semantic structures.
One common goal of probabilistic programming languages (PPLs) is to compute posterior probabilities for arbitrary
models and queries, given observed evidence, using a generic inference
engine.  Most PPL inference engines---even the
compiled ones---incur significant runtime interpretation overhead, especially for contingent and open-universe models. This
paper describes \name, a compiler for the BLOG PPL.
{\name}-generated code incorporates optimizations that eliminate
interpretation overhead, maintain dynamic dependencies efficiently,
and handle memory management for possible worlds of varying
sizes. Experiments comparing \name with other PPL engines on a
variety of inference problems demonstrate speedups ranging from 12x to
326x.
\end{abstract}

\section{Introduction}
Probabilistic programming languages (PPLs) aim to combine sufficient
expressive power for writing real-world probability models with
efficient, general-purpose inference algorithms that can answer
arbitrary queries with respect to those models. One underlying motive
is to relieve the user of the obligation to carry out machine learning
research and implement new algorithms for each problem that comes
along. Another is to support a wide range of cognitive functions in AI systems and to model those functions in humans.

General-purpose inference for PPLs is very challenging; they may
include unbounded numbers of discrete and continuous
variables, a rich library of distributions, and the ability to
describe uncertainty over functions, relations, and the existence and
identity of objects (so-called {\em open-universe} models).  Existing
PPL inference algorithms include likelihood weighting
(LW)~\cite{milch2005approximate}, parental Metropolis--Hastings
(PMH)~\cite{milch2006general,goodman2008church}, generalized Gibbs sampling
(Gibbs)~\cite{arora2010gibbs}, generalized sequential Monte Carlo~\cite{wood2014new}, Hamiltonian Monte Carlo (HMC)~\cite{sdt2014stan}, variational methods~\cite{minka2014infernet,kucukelbir2015automatic} and a form of approximate Bayesian
computation~\cite{mansinghka2013approximate}. While better
algorithms are certainly possible, our focus in this paper is on
achieving orders-of-magnitude improvement in the execution efficiency
of a given algorithmic process.

A PPL system takes a probabilistic program (PP) specifying a
probabilistic model as its input and performs inference to compute the
posterior distribution of a {\em query} given some observed {\em evidence}.
The inference process does not (in general) {\em execute}
the PP, but instead executes the steps of an inference algorithm (e.g.,~Gibbs) guided by the dependency structures implicit in the PP.  In
many PPL systems the PP exists as an internal data structure
consulted by the inference algorithm at each step
~\cite{pfeffer2001ibal,lunn2000winbugs,plummer2003jags,milch2005blog,pfeffer2009figaro}.
This process is in essence an {\em interpreter} for the PP, similar to
early Prolog systems that interpreted the logic program. Particularly
when running sampling algorithms that involve millions of repetitive
steps, the overhead can be enormous.  A natural solution is to produce model-specific compiled inference code, but, as we show in Sec.~\ref{sec:related}, existing compilers for general open-universe  models~\cite{wingate2011lightweight,yanggenerating,hur2014slicing,chaganty2013efficiently,nori2014r2} miss out on optimization opportunities and often produce inefficient inference code.

The paper analyzes the optimization opportunities for PPL compilers and describes the \name compiler, which takes as input a BLOG
program~\cite{milch2005blog} and one of three inference algorithms (LW, PMH, Gibbs) and
generates target code for answering queries. \name includes
three main contributions:
\begin{inparaenum}[(1)]
\item elimination of interpretative overhead by
joint analysis of the model structure and inference algorithm;
\item a dynamic
slice maintenance method (\optname) for incremental computation of the
current dependency structure as sampling proceeds; and
\item efficient
runtime memory management for maintaining the current-possible-world
data structure as the number of objects changes.
\end{inparaenum}
Comparisons
between \name and other PPLs on a variety of models demonstrate
speedups ranging from 12x to 326x, leading in some cases to
performance comparable to that of hand-built model-specific code. To
the extent possible, we also analyze the contributions of each
optimization technique to the overall speedup.

Although \name is developed for the BLOG language, the overall design and the choices of optimizations can be applied to other PPLs and may bring useful insights to similar AI systems for real-world applications.

\section{Existing PPL Compilation Approaches}
\label{sec:related}
In a general purpose programming language (e.g., C++), the compiler compiles exactly what the user writes (the program). By contrast, a PPL compiler essentially compiles the inference algorithm, which is written by the PPL developer, as applied to a PP. This means that different implementations of the same inference algorithm for the same PP result in completely different target code.

\begin{figure}[tb]
  \centering
    \includegraphics[width=0.48\textwidth]{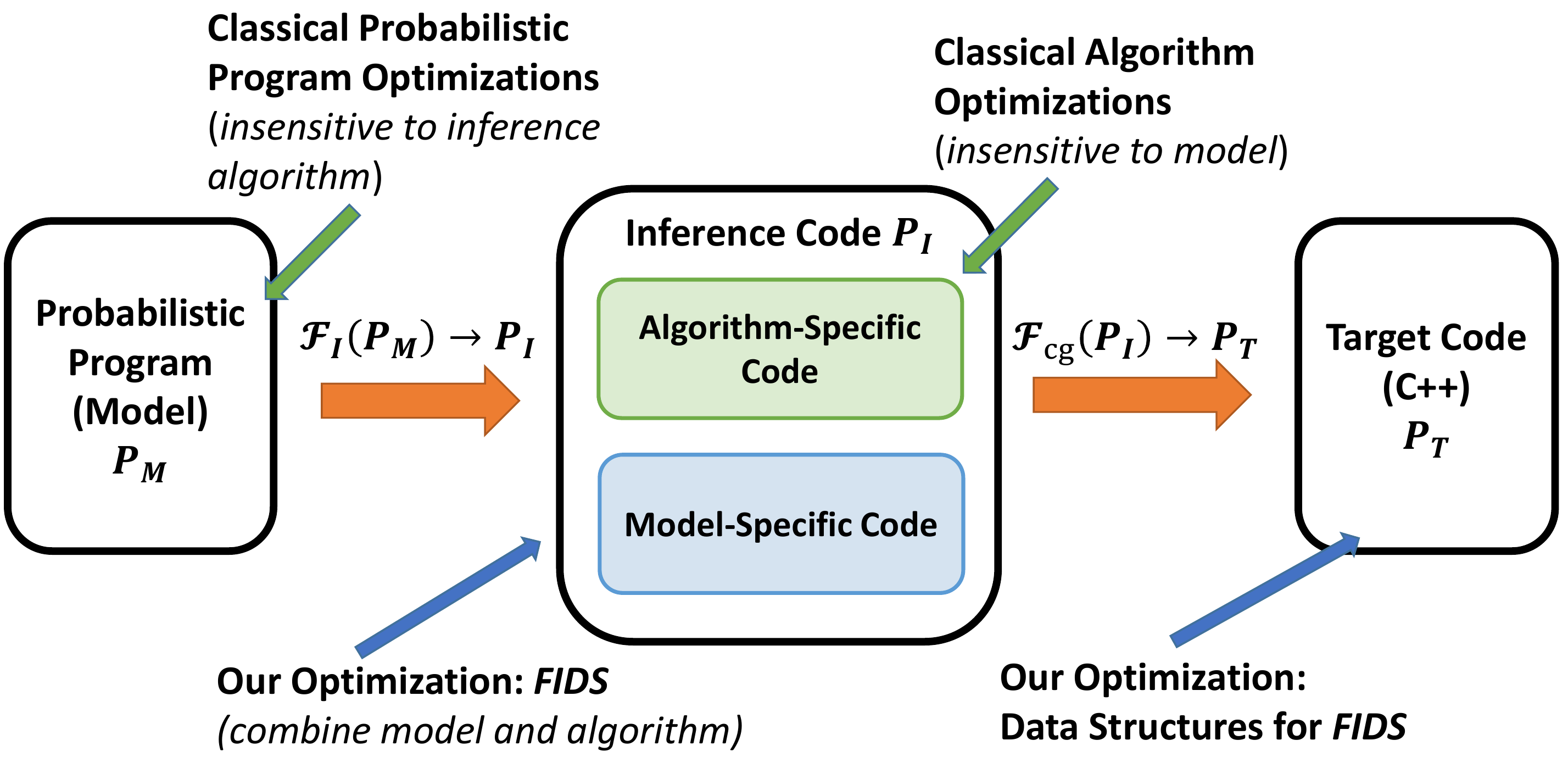}
    \caption{PPL compilers and optimization opportunities.}
    \label{fig:ppl-compile}
\end{figure}

As shown in Fig.~\ref{fig:ppl-compile}, a PPL compiler first produces an intermediate representation combining the inference algorithm ($I$) and the input model ($P_M$) as the \emph{inference code} ($P_I$), and then compiles $P_I$ to the \emph{target code} ($P_T$). \name focuses on optimizing the inference code $P_I$ and the target code $P_T$ given a fixed input model $P_M$.

Although there have been successful PPL compilers, these compilers are all designed for a restricted class of models with fixed dependency structures: see work by Tristan~\shortcite{NIPS2014_5531} and the Stan Development Team~\shortcite{sdt2014stan} for closed-world Bayes nets, Minka~\shortcite{minka2014infernet} for factor graphs, and Kazemi and Poole~\shortcite{kazemi2016knowledge} for Markov logic networks.

Church~\cite{goodman2008church} and its variants~\cite{wingate2011lightweight,yanggenerating,ritchie2015c3} provide a lightweight compilation framework for performing the Metropolis--Hastings algorithm (MH) over general open-universe probability models (OUPMs). However, these approaches (1) are based on an inefficient implementation of MH, which results in overhead in $P_I$, and (2) rely on inefficient data structures in the target code $P_T$.

For the first point, consider an MH iteration where we are proposing a new possible world $w'$ from $w$ accoding to the proposal $g(\cdot)$ by resampling random variable $X$ from $v$ to $v'$. The computation of acceptance ratio $\alpha$ follows
{
\begin{equation}\label{eq:naive-accept}
\alpha=\min\left(1,\frac{g(v'\to v)\Pr[w']}{g(v\to v')\Pr[w]}\right).
\end{equation}}
Since only $X$ is resampled, it is sufficient to compute $\alpha$ using merely the Markov blanket of $X$, which leads to a much simplified formula for $\alpha$ from Eq.(\ref{eq:naive-accept}) by cancelling terms in $\Pr[w]$ and $\Pr[w']$.
However, when applying MH for a contingent model, the Markov blanket of a random variable cannot be determined at compile time since the model dependencies vary during inference. This introduces tremendous runtime overhead for interpretive systems (e.g., BLOG, Figaro), which track all the dependencies at runtime even though typically only a tiny portion of the dependency structure may change per iteration.
Church simply avoids keeping track of dependencies and uses Eq.(\ref{eq:naive-accept}), including a potentially huge overhead for redundant probability computations.

For the second point, in order to track the existence of the variables in an open-universe model, similar to BLOG, Church also maintains a complicated dynamic string-based hashing scheme in the target code, which again causes interpretation overhead.

Lastly, techniques proposed by Hur \emph{et al.}~\shortcite{hur2014slicing}, Chaganty  \emph{et al.}~\shortcite{chaganty2013efficiently} and Nori \emph{et al.}~\shortcite{nori2014r2} primarily focus on optimizing $P_M$ by analyzing the static properties of the input PP, which are complementary to \name. 


\section{Background} 
This paper focuses on the BLOG language, although our approach also
applies to other PPLs with equivalent semantics~\cite{mcallester2008random,wu2014bfit}. Other PPLs can be also converted to BLOG via static single assignment form (SSA form) transformation~\cite{cytron1989efficient,hur2014slicing}.

\subsection{The BLOG Language}
The BLOG language~\cite{milch2005blog} defines probability measures over first-order (relational) possible worlds; in this sense it is a probabilistic analogue of first-order logic.
A BLOG program declares {\em types} for objects and defines distributions over their numbers, as well as defining distributions for the values of random functions applied to objects. {\em Observation statements} supply evidence, and a {\em query} statement specifies the posterior probability of interest. A random variable in BLOG
corresponds to the application of a random function to specific
objects in a possible world.
BLOG naturally supports open universe probability models (OUPMs) and context-specific dependencies.

\begin{figure}[tb]
\includegraphics[width=0.48\textwidth]{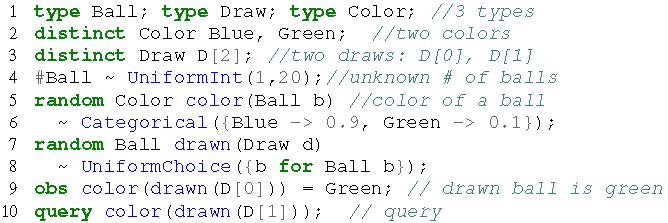}
\caption{The urn-ball model}\label{fig:blog-urn-ball}
\end{figure}

Fig.~\ref{fig:blog-urn-ball} demonstrates the open-universe urn-ball model. In this version the query asks for the color of the next random pick from an urn given the colors of balls drawn previously.
Line 4 is a \emph{number statement} stating that the \emph{number variable} $\#Ball$, corresponding to the total number of balls, is uniformly distributed between 1 and 20.
Lines 5--6 declare a random function $color(\cdot)$, applied to balls, picking \verb|Blue| or \verb|Green| with a biased probability.
Lines 7--8 state that each draw chooses a ball at random from the urn, with replacement.

\begin{figure}[tb]
\includegraphics[width=0.48\textwidth]{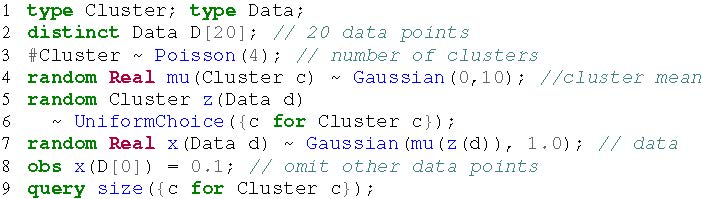}
\caption{The infinite Gaussian mixture model}\label{fig:infty-gmm}
\end{figure}

Fig.~\ref{fig:infty-gmm} describes another OUPM, the infinite Gaussian mixture model ($\infty$-GMM). The model includes an unknown number of clusters as stated in line 3, which is to be inferred from the data.

\subsection{Generic Inference Algorithms}
All PPLs need to infer the posterior distribution of a query given the observed evidence. The great majority of PPLs use Monte Carlo sampling methods such as likelihood weighting (LW) and Metropolis--Hastings MCMC (MH). LW samples unobserved random variables sequentially in topological order, conditioned on evidence and sampled values earlier in the sequence; each complete sample is weighted by the likelihood of the evidence variables appearing in the sequence. The MH algorithm is summarized in Alg.~\ref{alg:mh}.
\m denotes a BLOG model, \evi its evidence, \q a query, $N$ the number of samples, $w$ a possible world, $X(w)$ the value of random variable $X$ in $w$, $\Pr[X(w)|w_{{\bno}X}]$ denotes the conditional probability of $X$ in $w$, and $\Pr[w]$ denotes the likelihood of possible world $w$.
In particular, when the proposal distribution $g$ in Alg.\ref{alg:mh} samples a variable conditioned on its parents, the algorithm becomes the parental MH algorithm (PMH).
Our discussion in the paper focuses on LW and PMH but our proposed solution applies to other Monte Carlo methods as well.

\hide{
\begin{algorithm}[tb]
\caption{Likelihood-Weighting (LW)
\label{alg:lw}
}
\KwIn{\m, \evi, \q, $N$; {\bf Output}: histogram of samples $H$}
\For{$i\gets 1$ \textrm{to} $N$}{
  create empty possible world $W^i \leftarrow \emptyset$, $w_i \leftarrow 1$\;
	\ForEach{evidence $\texttt{obs }e_j=v$ in \evi}{
	    sample all the ancestors of $e_j$ and update $W_i$\;
	    $W_i(e_i)\gets v$; $w_i \leftarrow w_i \cdot \Pr[e_i|W_i]$\;
		\label{alg:lw:likeli}
	}
	sample \q and update $W_i$; $H \gets H+(W_i(Q),w_i)$\;
}
\end{algorithm}
}

\begin{algorithm}[tb]
\caption{Metropolis--Hastings algorithm (MH)
\label{alg:mh}
}
\KwIn{\m, \evi, \q, $N$; {\bf Output}: samples $H$}
initialize a possible world $w^{(0)}$ with \evi satisfied\;
\For{$i\gets 1$ \textrm{to} $N$}{
  randomly pick a variable $X$ from $w^{(i-1)}$ \;
  $w^{(i)}\gets w^{(i-1)}$ and $v\gets X(w^{(i-1)})$\;
  propose a value $v'$ for $X$ via proposal $g$\;
  $X(w^{(i)})\gets v'$ and ensure $w^{(i)}$ is self-supporting\;
  $\alpha\gets\min\left(1,\frac{g(v'\to v)\Pr[w^{(i)}]}{g(v\to v')\Pr[w^{(i-1)}]}\right)$ \label{eq:accept-rate}\;
  \lIf{$\mathrm{rand}(0,1) \ge \alpha$}{ $w^{(i)} \gets w^{(i-1)}$}
		$H \gets H+(\q(w^{(i)}))$\;
}
\end{algorithm}


\section{The Swift Compiler}
\name has the following design goals for handling open-universe probability models.
\begin{itemize}
    \item Provide a general framework to automatically and efficiently (1) track the exact Markov blanket for each random variable and (2) maintain the minimum set of random variables necessary to evaluate the query and the evidence (the dynamic slice~\cite{agrawal1990dynamic}, or equivalently the minimum self-supporting partial world~\cite{milch2005blog}).
    \item Provide efficient memory management and fast data structures for Monte Carlo sampling algorithms to avoid interpretive overhead at runtime.
\end{itemize}

For the first goal, we propose a novel framework, \optname (Framework for Incrementally updating Dynamic Slices), for generating the inference code ($P_I$ in Fig.~\ref{fig:ppl-compile}). For the second goal, we carefully choose data structures in the target C++ code ($P_T$).

For convenience, r.v.~is short for random variable. We demonstrate examples of compiled code in the following discussion: the inference code ($P_I$) generated by \optname in Sec.~\ref{sec:fids} is in pseudo-code while the target code ($P_T$) by \name shown in Sec.~\ref{sec:swift} is in C++. Due to limited space, we show the detailed transformation rules only for the adaptive contingency updating (ACU) technique and omit the others.

\subsection{Optimizations in FIDS}\label{sec:fids}
Our discussion in this section focuses on LW and PMH, although \optname can be also applied to other algorithms.
\optname includes three optimizations: dynamic backchaining (DB), adaptive contingency updating (ACU), and reference counting (RC). DB is applied to both LW and PMH while ACU and RC are specific to PMH.

\subsubsection{Dynamic backchaining}
\label{sec:opt-lw}
Dynamic backchaining (DB)~\cite{milch2005blog} constructs a dynamic slice incrementally in each LW iteration and samples only those variables necessary at runtime. DB in \name{} is an example of compiling lazy evaluation: in each iteration, DB backtracks from the evidence and query and samples an r.v.~only when its value is required during inference. 

For every r.v.~declaration \texttt{random T $X\sim C_X$;} in the input PP, \optname generates a \emph{getter} function \texttt{get{\ul}X()}, which is used to (1) sample a value for $X$ from its declaration $C_X$ and (2) memoize the sampled value for later references in the current iteration.
Whenever an r.v.~is referred to during inference, its getter function will be evoked.

Compiling DB is the fundamental technique for \name. The key insight is to replace dependency look-ups and method invocations from some internal PP data structure with direct machine address accessing and branching in the target executable file.

For example, the inference code for the getter function of $x(d)$ in the $\infty$-GMM model (Fig.~\ref{fig:infty-gmm}) is shown below.
\vspace{0.05cm}

\hide{
\begin{minted}[fontsize=\footnotesize]{cpp}
double get_x(Data d) {
  // some code memoizing the sampled value
  memoization;
  // if not sampled, sample a new value
  val = sample_gaussian(get_mu(get_z(d)),1);
  return val; }
\end{minted}
}
\noindent
\includegraphics{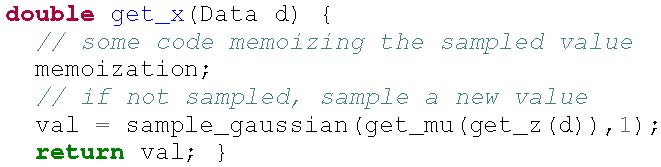}

\verb|memoization| denotes some pseudo-code snippet for performing memoization.  Since \verb|get_z|($\cdot$) is called before \verb|get_mu|($\cdot$), only those $mu(c)$ corresponding to non-empty clusters will be sampled. Notably, with the inference code above, no explicit dependency look-up is ever required at runtime to discover those non-empty clusters.

When sampling a number variable, we need to allocate memory for the associated variables. For example, in the urn-ball model (Fig.~\ref{fig:blog-urn-ball}), we need to allocate memory for $color(b)$ after sampling $\#Ball$. 
The corresponding generated inference code is shown below.
\vspace{0.1cm}

\hide{
\begin{minted}[fontsize=\footnotesize]{cpp}
int get_num_Ball() {
  // some code for memoization
  memoization;
  // sample a value
  val = sample_uniformInt(1,20);
  // some code allocating memory for color
  allocate_memory_color(val);
  return val; }
\end{minted}
}
\noindent
\includegraphics{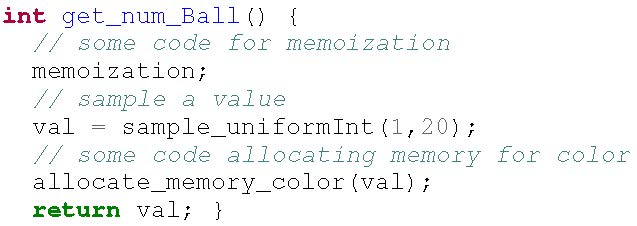}

{\footnotesize \verb|allocate_memory_color(val)|}
denotes some pseudo-code segment for allocating \verb|val| chunks of memory for the values of $color(b)$.

\subsubsection{Reference Counting}
Reference counting (RC) generalize the idea of DB to incrementally maintain the dynamic slice in PMH. RC is an efficient compilation strategy for the interpretive BLOG to dynamically maintain references to variables and exclude those without any references from the current possible world with minimal runtime overhead. RC is also similar to dynamic garbage collection in programming language community.

For an r.v.~being tracked, say $X$, RC maintains a reference counter $\texttt{cnt(X)}$ defined by $\texttt{cnt(X)}=|Ch_w(X)|$. $Ch_w(X)$ denote the children of $X$ in the possible world $w$. When $\texttt{cnt(X)}$ becomes zero, $X$ is removed from $w$; when $\texttt{cnt(X)}$ become positive, $X$ is instantiated and added back to $w$. This procedure is performed recursively.

Tracking references to every r.v.~might cause unnecessary overhead. For example, for classical Bayes nets, RC never excludes any variables. Hence, as a trade-off, \name only counts references in open-universe models, particularly, to those variables associated with number variables (e.g., $color(b)$ in the urn-ball model).

Take the urn-ball model as an example. When resampling $drawn(d)$ and accepting the proposed value $v$, the generated code for accepting the proposal will be\\
\hide{
\begin{minted}[fontsize=\footnotesize]{cpp}
void inc_cnt(X) {
  if (cnt(X) == 0) W.add(X);
  cnt(X) = cnt(X) + 1; }
void dec_cnt(X) {
  cnt(X) = cnt(X) - 1;
  if (cnt(X) == 0) W.remove(X); }
void accept_value_drawn(Draw d, Ball v) {
// code for updating dependencies omitted
  dec_cnt(color(val_drawn(d)));
  val_drawn(d) = v;
  inc_cnt(color(v)); }
\end{minted}
}
\includegraphics{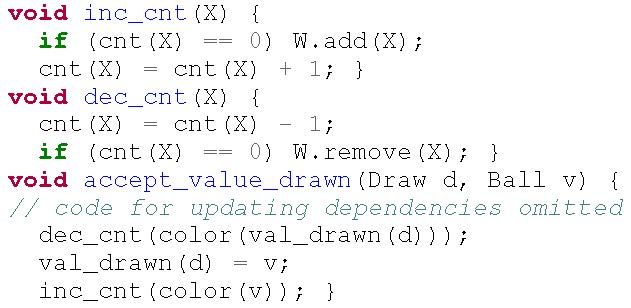}

The function {\footnotesize \verb|inc_cnt(X)|} and {\footnotesize \verb|dec_cnt(X)|} update the references to $X$. The function {\footnotesize \verb|accept_value_drawn()|} is specialized to r.v.~$drawn(d)$: \name analyzes the input program and generates specialized codes for different variables.

\subsubsection{Adaptive contingency updating}

The goal of ACU is to incrementally maintain the Markov Blanket for every r.v.~with minimal efforts.

$C_X$ denotes the declaration of r.v.~$X$ in the input PP;
$Par_w(X)$ denote the parents of $X$ in the possible world $w$; $w[X\gets v]$ denotes the new possible world derived from $w$ by only changing the value of $X$ to $v$. Note that deriving $w[X\gets v]$ may require instantiating new variables not existing in $w$ due to dependency changes.

Computing $Par_w(X)$ for $X$ is straightforward by executing its generation process, $C_X$, within $w$, which is often inexpensive. Hence, the principal challenge for maintaining the Markov blanket is to efficiently maintain a \emph{children set} $\verb|Ch(X)|$ for each r.v.~$X$ with the aim of keeping identical to the true set $Ch_w(X)$ for the current possible world $w$.

With a proposed possible world (PW) $w'=w[X\gets v]$, it is easy to add all missing dependencies from a particular r.v.~$U$. That is, for an r.v.~$U$ and every r.v.~$V\in Par_{w'}(U)$, since $U$ must be in $Ch_{w'}(V)$, we can keep $\verb|Ch(V)|$ up-to-date by adding $U$ to $\verb|Ch(V)|$ if $U\not\in \verb|Ch(V)|$. Therefore, the key step is tracking the set of variables, $\Delta(w[X\gets v])$, which will have a set of parents in $w[X\gets v]$ different from $w$.
$$
\begin{array}{l}
\Delta(W[X\gets v])=\{Y:Par_w(Y)\ne Par_{w[X\gets v]}(Y)\}\\
\end{array}
$$

Precisely computing $\Delta(w[X\gets v])$ is very expensive at runtime but computing an upper bound for $\Delta(w[X\gets v])$ does not influence the correctness of the inference code.  Therefore, we proceed to find an upper bound that holds for any value $v$. We call this over-approximation the \emph{contingent set} $Cont_w(X)$.

Note that for every r.v.~$U$ with dependency that changes in $w(X\gets v)$, $X$ must be reached in the condition of a control statement on the execution path of $C_U$ within $w$. This implies $\Delta(w[X\gets v])\subseteq Ch_w(X)$. One straightforward idea is set $Cont_w(X)=Ch_w(X)$. However, this leads to too much overhead. For example, $\Delta(\cdot)$ should be always empty for models with fixed dependencies.

Another approach is to first find out the set of \emph{switching variables} $S_X$ for every r.v.~$X$, which is defined by
$$
S_X=\{Y:\exists w,v\,\,Par_w(X)\ne Par_{w[Y\gets v]}(X) \},
$$
This brings an immediate advantage: $S_X$ can be statically computed at compile time. However, computing $S_X$ is NP-Complete\footnote{Since the declaration $C_X$ may contain arbitrary boolean formulas, one can reduce the 3-SAT problem to computing $S_X$.}.

Our solution is to derive the approximation $\widehat{S}_X$ by taking the union of free variables of if/case conditions, function arguments, and set expression conditions in $C_X$, and then set
$$
Cont_w(X)=\{Y:Y\in Ch_w(X)\land X\in \widehat{S}_Y\}.
$$
$Cont_w(X)$ is a function of the sampling variable $X$ and the current PW $w$. Likewise, for every r.v.~$X$, we maintain a runtime contingent set $\verb|Cont(X)|$ identical to the true set $Cont_w(X)$ under the current PW $w$. $\verb|Cont(X)|$ can be incrementally updated as well.

Back to the original focus of ACU, suppose we are adding new the dependencies in the new PW $w'=w[X\gets v]$. There are three steps to accomplish the goal: (1) enumerating all the variables $U\in \verb|Cont(X)|$, (2) for all $V\in Par_{w'}(U)$, add $U$ to $\verb|Ch(V)|$, and (3) for all $V\in Par_{w'}(U)\cap \widehat{S}_U$, add $U$ to $\verb|Cont(V)|$. These steps can be also repeated in a similar way to remove the vanished dependencies.

Take the $\infty$-GMM model (Fig.~\ref{fig:infty-gmm}) as an example , when resampling $z(d)$, we need to change the dependency of r.v.~$x(d)$ since $Cont_w(z(d))=\{x(d)\}$ for any $w$. The generated code for this process is shown below.
\hide{
\begin{minted}[fontsize=\footnotesize]{cpp}
void x(d)::add_to_Ch() {
  Cont(z(d)) += x(d);
  Ch(z(d)) += x(d);
  Ch(mu(z(d))) += x(d); }
void z(d)::accept_value(Cluster v) {
// accept the proposed value v for r.v. z(d)
// code for updating references omitted
  for (u in Cont(z(d))) u.del_from_Ch();
  val_z(d) = v;
  for (u in Cont(z(d))) u.add_to_Ch(); }
\end{minted}
}
\vspace{0.05cm}

\noindent
\includegraphics{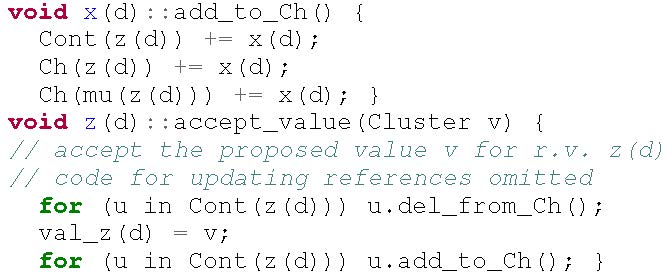}

We omit the code for {\footnotesize \verb|del_from_Ch()|} for conciseness, which is essentially the same as {\footnotesize \verb|add_to_Ch()|}.

The formal transformation rules for ACU are demonstrated in Fig.~\ref{fig:rule-ch}. $\mathcal{F}_{\texttt{c}}$ takes in a random variable declaration statement in the BLOG program and outputs the inference code containing methods \texttt{accept{\ul}value()} and \texttt{add{\ul}to{\ul}Ch()}. $\mathcal{F}_{\texttt{cc}}$ takes in a BLOG expression and generates the inference code inside method \texttt{add{\ul}to{\ul}Ch()}. It keeps track of already seen variables in $seen$, which makes sure that a variable will be added to the contingent set or the children set at most once. Code for removing variables can be generated similarly.

\begin{figure}[bt]
\begin{footnotesize}
$\mathcal{F}_{\texttt{c}}(\m = \texttt{random} \; \mathbf{T} \; \mathbf{Id}([\mathbf{T}\;\mathbf{Id}\;\texttt{,}]^*) \sim \mathbf{C}) =$ \\
\-\hspace{0.9cm} \lstinline[mathescape]!void $\mathbf{Id}$::add_to_Ch()! \\
\-\hspace{1cm} \lstinline[mathescape]!    {  $\mathcal{F}_{\texttt{cc}}(\mathbf{C}, \mathbf{Id}, \{ \} )$  }! \\
\-\hspace{1cm} \lstinline[mathescape]!void $\mathbf{Id}$::accept_value($\mathbf{T}$ v)! \\
\-\hspace{1cm} \lstinline[mathescape]! {  for(u in Cont($\mathbf{Id}$)) u.del_from_Ch();  ! \\
\-\hspace{1cm} \lstinline[mathescape]!    $\mathbf{Id}$ = v;! \\
\-\hspace{1cm} \lstinline[mathescape]!    for(u in Cont($\mathbf{Id}$)) u.add_to_Ch(); }! \\

$\mathcal{F}_{\texttt{cc}}(\mathbf{C} = \mathbf{Exp}, X, seen) = \mathcal{F}_{\texttt{cc}}(\mathbf{Exp}, X, seen)$ \\
$\mathcal{F}_{\texttt{cc}}(\mathbf{C} = \mathbf{Dist}(\mathbf{Exp}), X, seen) = \mathcal{F}_{\texttt{cc}}(\mathbf{Exp}, X, seen)$ \\
$\mathcal{F}_{\texttt{cc}}(\mathbf{C} = \texttt{if} \; (\mathbf{Exp}) \; \texttt{then} \; \mathbf{C}_1 \; \texttt{else} \; \mathbf{C}_2, X, seen) =$ \\
\-\hspace{1cm} $\mathcal{F}_{\texttt{cc}}(\mathbf{Exp}, X, seen)$; \\
\-\hspace{1cm} \lstinline[mathescape]!if ($\mathbf{Exp}$)! \\
\-\hspace{1cm} \lstinline[mathescape]!  {  $\mathcal{F}_{\texttt{cc}}(\mathbf{C}_1, X, seen \cup FV(\mathbf{Exp}))$;  }! \\
\-\hspace{1cm} \lstinline[mathescape]!else! \\
\-\hspace{1cm} \lstinline[mathescape]!  {  $\mathcal{F}_{\texttt{cc}}(\mathbf{C}_2, X, seen \cup FV(\mathbf{Exp}))$;  }! \\

$\mathcal{F}_{\texttt{cc}}(\mathbf{Exp}, X, seen) =$ \\
\-\hspace{0.2cm}/* \emph{emit a line of code for every r.v. U in the corresponding set} */\\
\-\hspace{1cm} $\forall \; U \in ((FV(\mathbf{Exp}) \cap \widehat{S}_X) \backslash seen):\,\,\texttt{Cont(U) += X}$; \\
\-\hspace{1cm} $\forall \; U \in (FV(\mathbf{Exp}) \backslash seen):\,\,\texttt{Ch(U) += X}$;
\end{footnotesize}
\caption{Transformation rules for outputting inference code with ACU. $FV(\mathbf{Expr})$ denote the free variables in $\mathbf{Expr}$. We assume the switching variables $\widehat{S}_X$ have already been computed. The rules for number statements and case statements are not shown for conciseness---they are analogous to the rules for random variables and if statements respectively.}
\label{fig:rule-ch}
\end{figure}

Since we maintain the exact children for each r.v.~with ACU, the computation of acceptance ratio $\alpha$ in Eq.~\ref{eq:naive-accept} for PMH can be simplified to

{
\begin{equation}\label{eq:accept-ratio-simple}
\min\left(1,\frac{|w|\Pr[X=v|w'_{{\bno}X}]\prod_{U \in Ch_{w'}(X)} \Pr[U(w') | w'_{{\bno}U}]}{|w'|\Pr[X=v'|w_{{\bno}X}]\prod_{V \in Ch_{w}(X)} \Pr[V(w) | w_{{\bno}V}]}\right)
\end{equation}
}
Here $|w|$ denotes the total number of random variables existing in $w$, which can be maintained via RC.

Finally, the computation time of ACU is strictly shorter than that of acceptance ratio $\alpha$ (Eq.~\ref{eq:accept-ratio-simple}).


\subsection{Implementation of Swift}\label{sec:swift}

\subsubsection{Lightweight memoization}\label{sec:impl-ds-memo}
Here we introduce the implementation details of the memoization code in the \emph{getter} function mentioned in section~\ref{sec:opt-lw}.

Objects will be converted to integers in the target code. \name analyzes the input PP and allocates static memory for memoization in the target code. For open-universe models where the number of random variables are unknown, the \emph{dynamic table} data structure (e.g., \texttt{vector} in C++) is used to reduce the amount of dynamic memory allocations: we only increase the length of the array when the number becomes larger than the capacity.

One potential weakness of directly applying a dynamic table is that, for models with multiple nested number statements (e.g. the aircraft tracking model with multiple aircrafts and blips for each one), the memory consumption can be large. In \name, the user can force the target code to clear all the unused memory every fixed number of iterations via a compilation option, which is turned off by default.

The following code demonstrates the target code for the number variable $\#Ball$ and its associated $color(b)$'s in the urn-ball model (Fig.~\ref{fig:blog-urn-ball}) where \texttt{iter} is the current iteration number.
\vspace{0.1cm}

\noindent
\includegraphics{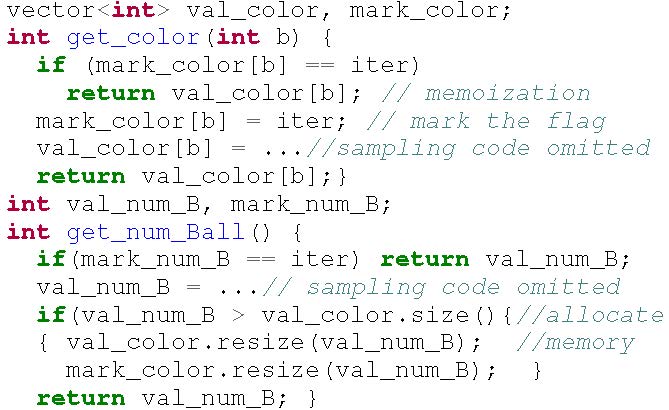}

Note that when generating a new PW, by simply increasing the counter \texttt{iter}, all the memoization flags (e.g., \texttt{mark\underline{ }color} for $color(\cdot)$) will be automatically cleared.

Lastly, in PMH, we need to randomly select an r.v.~to sample per iteration. In the target C++ code, this process is accomplished via polymorphism by declaring an abstract class with a \texttt{resample()} method and a derived class for every r.v.~in the PP. An example code snippet for the $\infty$-GMM model is shown below.

\hide{
\begin{minted}[fontsize=\footnotesize]{cpp}
class MH_OBJ {public: //abstract class
  virtual void resample()=0;//resample step
  virtual void add_to_Ch()=0; //for ACU
  virtual void accept_value()=0;//for ACU
  virtual double get_likeli()=0;
}; // some methods omitted
// maintain all the r.v.s in the current PW
std::vector<MH_OBJ*> active_vars;
// derived classes for r.v.s in the PP
class Var_mu:public MH_OBJ{public://for mu(c)
  double val; // value for the var
  double cached_val; //cache for proposal
  int mark, cache_mark; // flags
  std::set<MH_OBJ*> Ch, Cont; //sets for ACU
  double getval(){...}; //sample & memoize
  double getcache(){...};//generate proposal
  ... }; // some methods omitted
std::vector<Var_mu*> mu; // dynamic table
class Var_z:public MH_OBJ{ ... };//for z(d)
Var_z* z[20]; // fixed-size array
\end{minted}
}
\vspace{0.02cm}
\noindent
\includegraphics{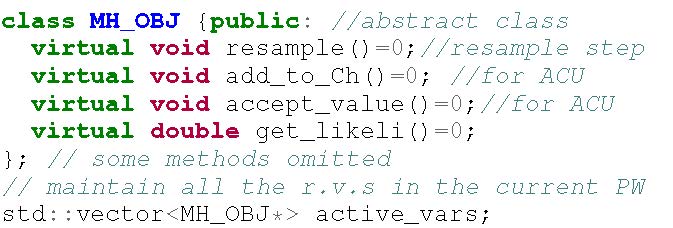}
\includegraphics{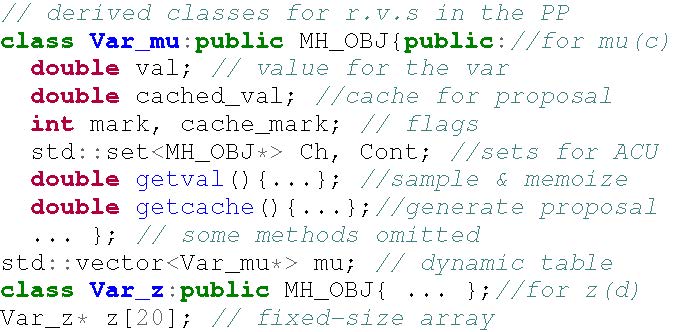}

\subsubsection{Efficient proposal manipulation}\label{sec:impl-ds-prop}

Although one PMH iteration only samples a single variable, generating a new PW may still involve (de-)instantiating an arbitrary number of random variables due to RC or sampling a number variable. The general approach commonly adopted by many PPL systems is to construct the proposed possible world in dynamically allocated memory and then copy it to the current world~\cite{milch2005blog,wingate2011lightweight}, which suffers from significant overhead.

In order to generate and accept new PWs in PMH with negligible memory management overhead, we extend the lightweight memoization to manipulate proposals: \name statically allocates an extra memoized cache for each random variable, which is dedicated to storing the proposed value for that variable. During resampling, all the intermediate results are stored in the cache. When accepting the proposal, the proposed value is directly loaded from the cache; when rejection, no action is needed due to our memoization mechanism.

Here is an example target code fragment for $mu(c)$ in the $\infty$-GMM model.
\texttt{proposed{\ul}vars} is an array storing all the variables to be updated if the proposal gets accepted.
\vspace{0.08cm}

\noindent
\includegraphics{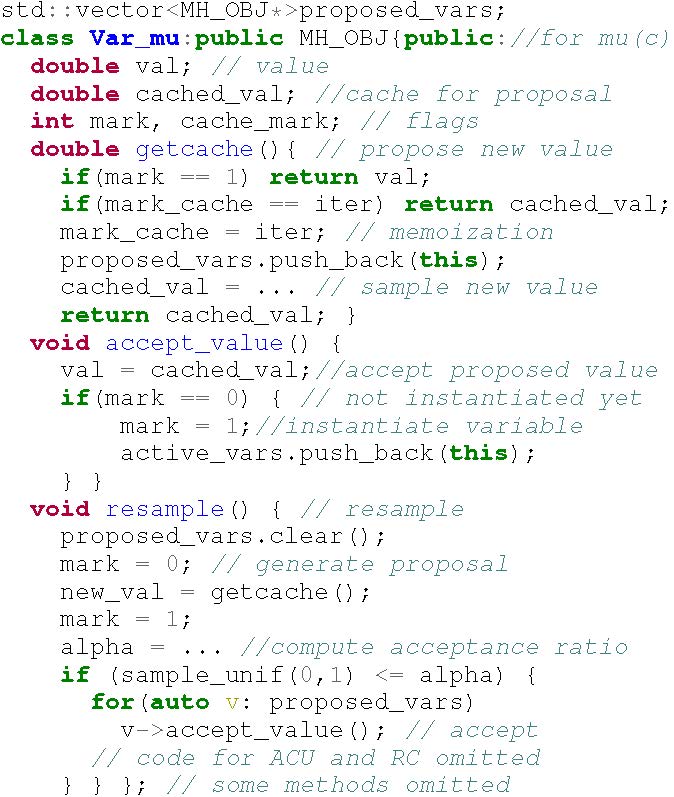}
\hide{
\begin{small}
\begin{minted}{cpp}
std::vector<MH_OBJ*>proposed_vars;
class Var_mu:public MH_OBJ{public://for mu(c)
  double val; // value
  double cached_val; //cache for proposal
  int mark, cache_mark; // flags
  double getcache(){ // propose new value
    if(mark == 1) return val;
    if(mark_cache == iter) return cached_val;
    mark_cache = iter; // memoization
    proposed_vars.push_back(this);
    cached_val = ... // sample new value
    return cached_val; }
  void accept_value() {
    val = cached_val;//accept proposed value
    if(mark == 0) { // not instantiated yet
        mark = 1;//instantiate variable
        active_vars.push_back(this);
    } }
  void resample() { // resample
    proposed_vars.clear();
    mark = 0; // generate proposal
    new_val = getcache();
    mark = 1;
    alpha = ... //compute acceptance ratio
    if (sample_unif(0,1) <= alpha) {
      for(auto v: proposed_vars)
        v->accept_value(); // accept
      // code for ACU and RC omitted
    } } }; // some methods omitted
\end{minted}
\end{small}
}

\subsubsection{Supporting new algorithms}\label{sec:impl-new-algo}
We demonstrate that \optname can be applied to new algorithms by implementing a translator for the Gibbs sampling~\cite{arora2010gibbs} (Gibbs) in \name.

Gibbs is a variant of MH with the proposal distribution $g(\cdot)$ set to be the posterior distribution. The acceptance ratio $\alpha$ is always 1 while the disadvantage is that it is only possible to explicitly construct the posterior distribution with conjugate priors. In Gibbs, the proposal distribution is still constructed from the Markov blanket, which again requires maintaining the children set. Hence, \optname can be fully utilized.

However, different conjugate priors yield different forms of posterior distributions. In order to support a variety of proposal distributions, we need to (1) implement a conjugacy checker to check the form of posterior distribution for each r.v.~in the PP at compile time (the checker has nothing to do with \optname); (2) implement a posterior sampler for every conjugate prior in the runtime library of \name.


\section{Experiments}
In the experiments, \name generates target code in C++ with C++ standard \texttt{<random>} library for random number generation and the armadillo package~\cite{sanderson2010armadillo} for matrix computation.
The baseline systems include BLOG (version 0.9.1), Figaro (version 3.3.0), Church (webchurch), Infer.NET (version 2.6), BUGS (winBUGS 1.4.3) and Stan (cmdStan 2.6.0).

\subsection{Benchmark models}

We collect a set of benchmark models\footnote{Most models are simplified to ensure that all the benchmark systems can produce an output in reasonably short running time. All of these models can be enlarged to handle more data without adding extra lines of BLOG code.} which exhibit various capabilities of a PPL (Tab.~\ref{tab:models}), including the burglary model (Bur), the hurricane model (Hur), the urn-ball model (U-B($x$,$y$) denotes the urn-ball model with at most $x$ balls and $y$ draws), the TrueSkill model~\cite{herbrich2007trueskill} (T-K), 1-dimensional Gaussian mixture model (GMM, 100 points with 4 clusters) and the infinite GMM model ($\infty$-GMM).
We also include a real-world dataset: handwritten digits~\cite{lecun1998gradient} using the PPCA model~\cite{tipping1999probabilistic}.
All the models can be downloaded from the GitHub repository of BLOG.

\begin{table}[tb]
\centering
{\footnotesize
\begin{tabular}{cccccccc}
\toprule
 &Bur &Hur	&U-B&T-K &GMM &{\scriptsize$\infty$-}GMM&PPCA \\
\midrule
R &&&&\checkmark&\checkmark&\checkmark&\checkmark \\
CT&	&\checkmark	&\checkmark&&\checkmark&\checkmark& \\
CC&	&\checkmark	&&&&& \\
OU&	&&\checkmark&&&\checkmark& \\
CG&\checkmark	&\checkmark&\checkmark&&\checkmark&\checkmark&\checkmark \\
\bottomrule
\end{tabular}
}
\caption{Features of the benchmark models. R: continuous scalar or vector variables. CT: context-specific dependencies (contingency). CC: cyclic dependencies in the PP (while in any particular PW the dependency structure remains acyclic). OU: open-universe. CG: conjugate priors.}
\label{tab:models}
\end{table}

\subsection{Speedup by \optname within \name}
We first evaluate the speedup promoted by each of the three optimizations in \optname individually.

\subsubsection{Dynamic Backchaining for LW}
\label{sec:expr-DB}
We compare the running time of the following versions of code: (1) the code generated by \name (``Swift''); (2) the modified compiled code with DB manually turned off (``No DB''), which sequentially samples all the variables in the PP; (3) the hand-optimized code without unnecessary memoization or recursive function calls (``Hand-Opt'').

We measure the running time for all the 3 versions and the number of calls to the random number generator for ``Swift'' and ``No DB''. The result is concluded in Tab.~\ref{tab:lw}.

\begin{table}[bt]
  \centering
{\footnotesize
  \begin{tabular}{|c|c|c|c|c|}
    \hline
    \textbf{Alg.} & \textbf{Bur}& \textbf{Hur}& \textbf{U-B(20,2)}& \textbf{U-B(20,8)}\\
    \hline
    \multicolumn{5}{|c|}{\textbf{Running Time (s): Swift v.s. No DB}}\\
    \hline
    No DB& 0.768&1.288&2.952&5.040\\
    \hline
    Swift & 0.782&1.115&1.755&4.601\\
    \hline
    \textbf{\emph{Speedup}}&0.982&\textbf{1.155}&\textbf{1.682}&\textbf{1.10} \\
    \hline
    \multicolumn{5}{|c|}{\textbf{Running Time (s): Swift v.s. Hand-Optimized}}\\
    \hline
    Hand-Opt & 0.768 &1.099&1.723&4.492\\
    \hline
    Swift & 0.782&1.115&1.755&4.601\\
    \hline
    \emph{Overhead}&1.8\%&1.4\%&1.8\%&2.4\% \\
    \hline
    \multicolumn{5}{|c|}{\textbf{Calls to Random Number Generator}}\\
    \hline
    No DB & $3*10^7$&$3*10^7$&$1.35*10^8$&$1.95*10^8$\\
    \hline
    Swift & $3*10^7$&$2.0*10^7$&$4.82*10^7$&$1.42*10^8$\\
    \hline
    \emph{Calls Saved} & 0\% & 33.3\% &64.3\%&27.1\%\\
    \hline
  \end{tabular}
  }
  \caption{Performance on LW with $10^7$ samples for Swift, the version without DB and the hand-optimized version}
  \label{tab:lw}
\end{table}

The overhead due to memoization compared against the hand-optimized code is less than $2.4\%$. We can further notice that the speedup is  proportional to the number of calls saved.

\subsubsection{Reference Counting for PMH}
RC only applies to open-universe models in \name. Hence we focus on the urn-ball model with various model parameters.
The urn-ball model represents a common model structure appearing in many applications with open-universe uncertainty. This experiment reveals the potential speedup by RC for real-world problems with similar structures.
RC achieves greater speedup when the number of balls and the number of observations become larger.

\begin{figure}[tb]
\centering
\includegraphics[width=0.35\textwidth]{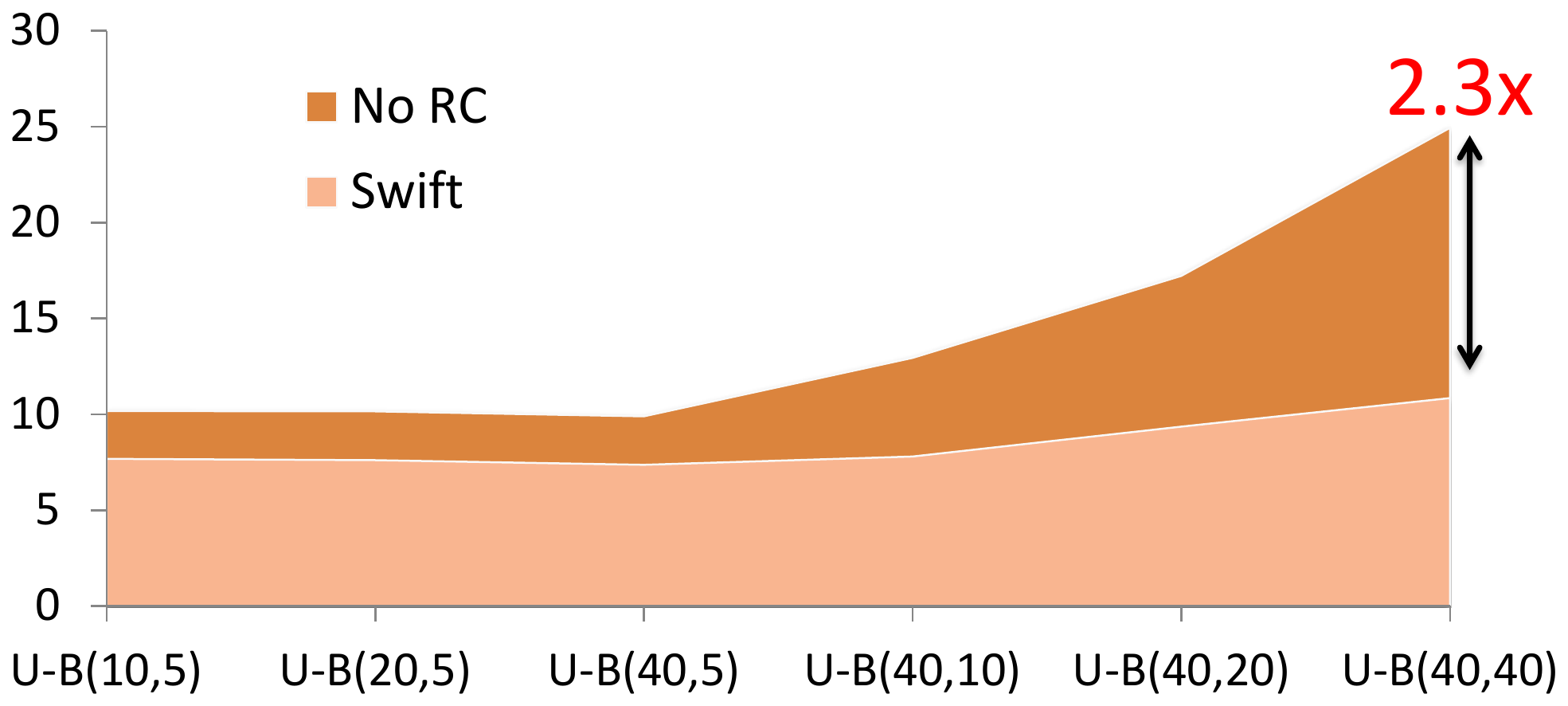}
\caption{Running time (s) of PMH in \name with and without RC on urn-ball models for 20 million samples.
\label{fig:mh-ref-time}}
\end{figure}

We compare the code produced by \name with RC (``Swift'') and RC manually turned off (``No RC'').  For ``No RC'', we traverse the whole dependency structure and reconstruct the dynamic slice every iteration.
Fig.~\ref{fig:mh-ref-time} shows the running time of both versions. RC is indeed effective -- leading to up to 2.3x speedup.

\subsubsection{Adaptive Contingency Updating for PMH}
We compare the code generated by \name with ACU (``Swift'') and the manually modified version without ACU (``Static Dep'') and measure the number of calls to the likelihood function in Tab.~\ref{tab:mh-acu-time}.

The version without ACU (``Static Dep'') demonstrates the best efficiency that can be achieved via compile-time analysis without maintaining dependencies at runtime. This version statically computes an upper bound of the children set by $\widehat{Ch}(X)=\{Y: X\textrm{ appears in }C_Y\}$ and again uses Eq.(\ref{eq:accept-ratio-simple}) to compute the acceptance ratio. Thus, for models with fixed dependencies, ``Static Dep'' should be faster than ``Swift''.

In Tab.~\ref{tab:mh-acu-time}, ``Swift'' is up to 1.7x faster than ``Static Dep'', thanks to up to $81\%$ reduction of likelihood function evaluations.
Note that for the model with fixed dependencies (Bur), the overhead by ACU is almost negligible: $0.2\%$ due to traversing the hashmap to access to child variables.

\begin{table}[tb]
  \centering
{\footnotesize
  \begin{tabular}{|c|c|c|c|c|}
    \hline
    \textbf{Alg.} & \textbf{Bur}& \textbf{Hur}&\textbf{U-B\scriptsize{(20,10)}}&\textbf{U-B\scriptsize{(40,20)}}\\
    \hline
    \multicolumn{5}{|c|}{\textbf{Running Time (s): Swift v.s. Static Dep}}\\
    \hline
    Static Dep& 0.986&1.642&4.433&7.722\\
    \hline
    Swift &0.988&1.492&3.891&4.514\\
    \hline
    \textbf{\emph{Speedup}} & \textbf{0.998} &\textbf{1.100}&\textbf{1.139}&\textbf{1.711}\\
    \hline
    \multicolumn{5}{|c|}{\textbf{Calls to Likelihood Functions}}\\
    \hline
    Static Dep& $1.8*10^7$ &$2.7*10^7$&$1.5*10^8$&$3.0*10^8$\\
    \hline
    Swift& $1.8*10^7$ &$1.7*10^7$&$4.1*10^7$&$5.5*10^7$\\
    \hline
    \emph{Calls Saved}& \textbf{0\%} &\textbf{37.6\%}&\textbf{72.8\%}&\textbf{81.7\%}\\
    \hline
  \end{tabular}
  }
  \caption{Performance of PMH in Swift and the version utilizing static dependencies on benchmark models.}
  \label{tab:mh-acu-time}
\end{table}

\subsection{\name against other PPL systems}

We compare \name with other systems using LW, PMH and Gibbs respectively on the benchmark models. The running time is presented in Tab.~\ref{tab:rt-all}. The speedup is measured between \name and the fastest one among the rest. An empty entry means that the corresponding PPL fails to perform inference on that model. Though these PPLs have different host languages (C++, Java, Scala, etc.), the performance difference resulting from host languages is within a factor of 2\footnote{\url{http://benchmarksgame.alioth.debian.org/}}.

\begin{table}[bt]
  \centering
  {\footnotesize
  \begin{tabular}{|c|c|c|c|c|c|c|}
    \hline
    \textbf{PPL} & \textbf{Bur}& \textbf{Hur}& \textbf{\stackanchor{U-B}{\scriptsize{(20,10)}}}& \textbf{T-K}&\textbf{GMM}&\textbf{\scriptsize{$\infty$-}GMM}\\
    \hline
    \multicolumn{7}{|c|}{\textbf{LW with $10^6$ samples}}\\
    \hline
    Church&9.6&22.9&179&57.4&1627&1038\\
    \hline
    Figaro&15.8&24.7&176&48.6&997&235\\
    \hline
    BLOG&8.4&11.9&189&49.5&998&261\\
    \hline
    \name&0.04&0.08&0.54&0.24&6.8&1.1\\
    \hline
    \textbf{\emph{Speedup}}&\textbf{196}&\textbf{145}&\textbf{326}&\textbf{202}&\textbf{147}&\textbf{214} \\
    \hline
    \multicolumn{7}{|c|}{\textbf{PMH with $10^6$ samples}}\\
    \hline
    Church&12.7&25.2&246&173&3703&1057\\
    \hline
    Figaro&10.6&-&59.6&17.4&151&62.2\\
    \hline
    BLOG&6.7&18.5&30.4&38.7&72.5&68.3\\
    \hline
    \name&0.11&0.15&0.4&0.32&0.76&0.60\\
    \hline
    \textbf{\emph{Speedup}}&\textbf{61}&\textbf{121}&\textbf{75}&\textbf{54}&\textbf{95}&\textbf{103} \\
    \hline
    \multicolumn{7}{|c|}{\textbf{Gibbs with $10^6$ samples}}\\
    \hline
    BUGS&87.7&-&-&-&84.4&-\\
    \hline
    Infer.NET&1.5&-&-&-&77.8&-\\
    \hline
    \name&0.12&0.19&0.34&-&0.42&0.86\\
    \hline
    \textbf{\emph{Speedup}}&\textbf{12}&\textbf{$\infty$}&\textbf{$\infty$}&-&\textbf{185}&\textbf{$\infty$} \\
    \hline
  \end{tabular}
  }
  \caption{Running time (s) of \name and other PPLs on the benchmark models using LW, PMH and Gibbs.}
  \label{tab:rt-all}
\end{table}

\subsection{Experiment on real-world dataset}
We use \name with Gibbs to handle the probabilistic principal component analysis (PPCA)~\cite{tipping1999probabilistic}, with real-world data: 5958 images of digit ``2'' from the hand-written digits dataset (MNIST)~\cite{lecun1998gradient}. We compute 10 principal components for the digits. The training and testing sets include 5958 and 1032 images respectively, each with 28x28 pixels and pixel value rescaled to $[0,1]$.

\hide{
PPCA is originally proposed
in \cite{tipping1999probabilistic} where each observation  $y_i\sim\mathcal{N}(Ax_i+\mu,\sigma^2 I)$,
where $A$ is a matrix with $K$ columns, $\mu$ is the mean vector, and $x_i$ is a coefficient vector associated with each data. All the entries of $A$,
$\mu$ and $x_i$ have independent Gaussian priors.

We use a subset of MNIST data
set~\cite{lecun1998gradient} for evaluation (corresponding to the digit ``2''). The training and testing sets include 5958 and 1032 images respectively, each with 28x28 pixels.
The pixel values are rescaled to the range $[0, 1]$. $K$ is set to 10.
}

Since most benchmark PPL systems are too slow to handle this amount of data, we compare \name against other two scalable PPLs, Infer.NET and Stan, on this model. Both PPLs have compiled inference and are widely used for real-world applications.
Stan uses HMC as its inference algorithm. For Infer.NET, we select variational message passing algorithm (VMP), which is Infer.NET's primary focus. 
Note that HMC and VMP are usually favored for fast convergence.

\begin{figure}[tb]
  \centering
    \includegraphics[width=0.45\textwidth]{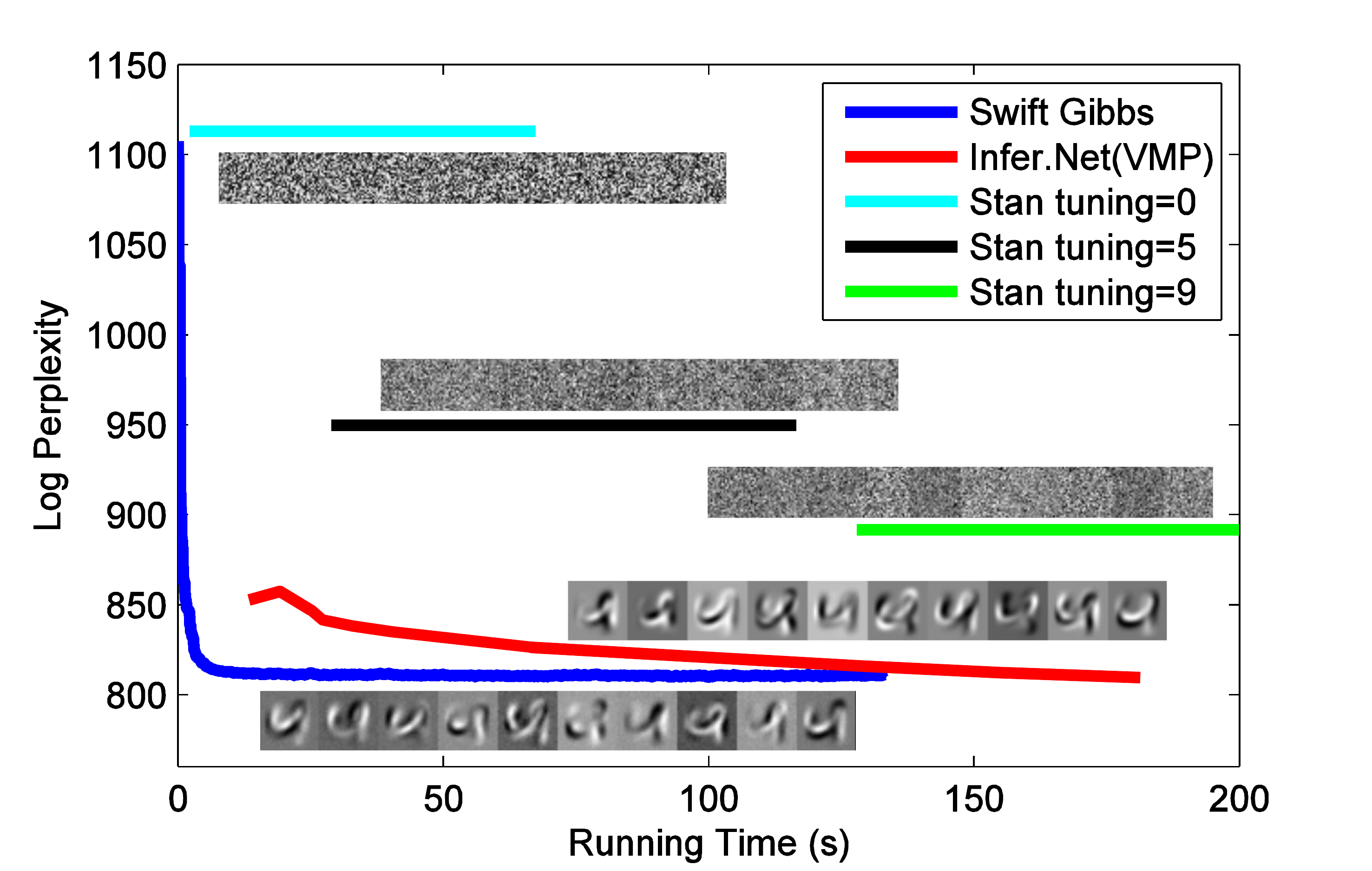}
    \caption{Log-perplexity w.r.t running time(s) on the PPCA model with visualized principal components. \name converges faster. }
    \label{fig:Swift-Stan-PPCA}
\end{figure}

Stan requires a tuning process before it can produce
samples. We ran Stan with 0, 5 and 9 tuning steps respectively\footnote{We also ran Stan with 50 and 100 tuning steps, which took more than 3 days to finish 130 iterations (including tuning).
However, the results are almost the same as that with 9 tuning steps.}. We measure the perplexity of the generated samples over test images w.r.t. the running time in Fig.~\ref{fig:Swift-Stan-PPCA}, where we also visualize the produced principal components. For Infer.NET, we consider the mean of the approximation distribution.

\name quickly generates visually meaningful outputs with around $10^5$ Gibbs iterations in 5 seconds. Infer.NET takes 13.4 seconds to finish the first iteration\footnote{Gibbs by \name samples a single variable while Infer.NET processes all the variables per iteration.} and converges to a result with the same perplexity after 25 iterations and 150 seconds. The overall convergence of Gibbs w.r.t.~the running time significantly benefits from the speedup by \name.

For Stan, its no-U-turn sampler~\cite{homan2014no}
suffers from significant parameter tuning issues. We also tried to manually tune the parameters, which does not help much. Nevertheless, Stan does work for the simplified PPCA model with 1-dimensional data.  Although further investigation is still needed, we conjecture that Stan is very sensitive to its parameters in high-dimensional cases. Lastly, this experiment also suggests that those parameter-robust inference engines, such as \name with Gibbs and Infer.NET with VMP, would be preferred at practice when possible.
\hide{
over the testing images from MNIST dataset. The perplexity w.r.t. the running time for Swift and Stan are shown in Figure \ref{fig:Swift-Stan-PPCA} with the produced principal components visualized.
We also ran Stan with 50 and 100 tuning steps (not shown in the figure).
However, the perplexity of samples with 50 and 100 tuning iterations are almost the same as those with 9 tunning iterations.
}

\section{Conclusion and Future Work} 

We have developed \name, a PPL compiler that generates model-specific target code for performing inference. Swift uses a dynamic slicing framework for open-universe models to incrementally maintain the dependency structure at runtime. In addition, we carefully design data structures in the target code to avoid dynamic memory allocations when possible. Our experiments show that Swift achieves orders of magnitudes speedup against other PPL engines.

The next step for \name is to support more algorithms, such as SMC~\cite{wood2014new}, as well as more samplers, such as the block sampler for (nearly) deterministic dependencies~\cite{li2013dynamic}. Although \optname is a general framework that can be naturally extended to these cases, there are still implementation details to be studied.

Another direction is partial evaluation, which allows the compiler to reason about the input program to simplify it. Shah \emph{et al.}~\shortcite{rohin2016simpl} proposes a partial evaluation framework for the inference code ($P_I$ in Fig.~\ref{fig:ppl-compile}). It is very interesting to extend this work to the whole \name pipeline.


\section*{Acknowledgement}
We would like to thank our anonymous reviewers, as
well as Rohin Shah for valuable discussions.
This work is supported by the DARPA PPAML program, contract FA8750-14-C-0011.

{\small
\bibliographystyle{named}
\bibliography{ijcai16}
}

\end{document}